\pdfoutput=1
\documentclass[11pt]{article}

\usepackage[]{naacl2021}

\usepackage{times}
\usepackage{latexsym}
\usepackage{graphicx}
\usepackage{tabularx}
\usepackage{amsfonts}
\usepackage{multirow} 
\usepackage[T1]{fontenc}
\usepackage[utf8]{inputenc}

\usepackage{microtype}

\title{BED: Bi-Encoder-Decoder Model for Canonical Relation Extraction}

\author{Nantao Zheng$^{1}$, Siyu Long$^{1}$, Xinyu Dai$^{1}$ \\
  $^{1}$National Key Laboratory for Novel Software Technology, Nanjing University \\
  {\tt {\{zhengnt,longsy\}}@smail.nju.edu.cn}\\
  {\tt daixinyu@nju.edu.cn}
}

\begin{document} 
\maketitle
\begin{abstract}
% Canonical relation extraction (CRE) aims to extract canonical relational triples from given sentences, where the triple elements (entity pairs and their relationship) are mapped to the existing knowledge base. Early studies in CRE cast the extraction process into several sub-processes, which suffer from the error propagation problem. Recently, to alleviate this problem, methods based on the encoder-decoder architecture, which directly translate sentences into triples, are proposed and achieve promising results. However, in these encoder-decoder based methods, entities are represented as indexes in the vocabulary, lacking the explicit process of utilizing the entity semantics. Moreover, due to the vocabulary size is fixed, when facing novel entities, costly retraining is required to obtain their representations. In this paper, we propose a novel framework, Bi-Encoder-Decoder (BED), to solve the above issues. Specifically, to explicitly utilize the entity semantics, an entity encoder is adopted to directly encode the entity name. For novel entities, their representation can be easily generated by using the previously trained entity encoder. Experimental results on two datasets show that the proposed framework achieves a significant performance improvement over the previous state-of-the-art and handle novel entities well without retraining.

Canonical relation extraction aims to extract relational triples from sentences, where the triple elements (entity pairs and their relationship) are mapped to the knowledge base. Recently, methods based on the encoder-decoder architecture are proposed and achieve promising results. 
% However, in these encoder-decoder based methods, entities are represented as indexes in the vocabulary, lacking explicit modeling of entity semantic representations. 
% However, these methods assign each entity an embedding which are learned independently during training.
% Moreover, following the close world assumption, they cannot represent novel entities without a costly retraining.
However, these methods cannot well utilize the entity information, which is merely used as augmented training data. Moreover, they are incapable of representing novel entities, since no embeddings have been learned for them.
In this paper, we propose a novel framework, \textbf{B}i-\textbf{E}ncoder-\textbf{D}ecoder (\textbf{BED}), to solve the above issues. Specifically, to fully utilize entity information, we employ an encoder to encode semantics of this information, leading to high-quality entity representations.
% Naturally, the representations of novel entities can also be easily generated using the previously trained entity encoder. 
For novel entities, given a trained entity encoder, their representations  can be easily generated.
% For novel entities, their representations can be easily generated by the previously trained entity encoder.
Experimental results on two datasets show that, our method achieves a significant performance improvement over the previous state-of-the-art and handle novel entities well without retraining.

\end{abstract}

\section{Introduction}

Knowledge bases (KBs) such as Freebase \citep{bollacker2008freebase} and Wikidata \citep{vrandevcic2014wikidata} consist of a large number of relational facts, which are organized as $\langle$subject, relation, object$\rangle$. KBs have been widely applied in many tasks including question answering \citep{yih2015semantic} and recommender systems \citep{zhang2016collaborative}. 

Nonetheless, KBs are far from completion. Canonical relation extraction (CRE) \citep{trisedya2019neural} is a crucial task for KB enrichment. It aims to extract relational triples from natural language text, where triple elements are mapped to the KB. Table \ref{example} shows an example of CRE. Given the sentence ``\textit{Megan Manthey's hometown is Ferndale, Washington.}", we need to recognize two entities ``\textit{Megan Manthey}" and ``\textit{Ferndale}" whose corresponding entity IDs in Wikidata are ``Q6669593" and ``Q670897", and also their relationship ``place of birth" whose relation ID in Wikidata is ``P19".

\begin{table}[t]
\centering
\begin{tabular}{c}

\hline
\textbf{Input sentence:}\\
\hline
Megan Manthey's hometown is Ferndale,\\ Washington.\\
\hline
\textbf{Canonical relational triples:}\\
\hline
$\langle$Q6669593, P19, Q670897$\rangle$\\
\hline

\end{tabular}
\caption{Example of canonical relation extraction.}
\label{example}
\end{table}

Early studies in CRE take a pipeline approach. They first identify all possible entity mentions \citep{lample2016neural, strubell2017fast} in a sentence, then link each possible entity mention to a corresponding entity entry in the KB \citep{hoffart2011robust, kolitsas2018end}, and finally perform relation classification \citep{lin2016neural, wang2016relation} for each linkable entity pair. However, such methods suffer from the error propagation problem. To alleviate this problem, \citet{trisedya2019neural} propose encoder-decoder-based methods, which directly translate sentences into triples and achieve promising results by avoiding cascading errors.

% Despite their success, these approaches still leave much to be desired. Firstly, they assign an embedding vector to each entity and trained using labeled instances. Due to the data sparsity, there are only a few instances for per entity. Even with pre-training and adding pairs of $\langle$Entity-name, Entity-ID$\rangle$ into the training set, it is still hard to obtain high-quality entity representations, which hinders their performance. Moreover, they assume that all test entities appear during training. However, in practical scenarios, KBs evolve fast with out-of-knowledge-base (OOKB) entities added frequently \citep{dai2020inductively}. To represent OOKB entities, they need to retrain the model, which is extremely time-consuming.

Despite their success, these approaches still leave much to be desired. Firstly, in these methods, each entity is represented as an index in the vocabulary and assigned an embedding vector which is trained using the sentence-triples pairs. To better align entity names and entity IDs, they use  Even with auxiliary pre-training and additional training data (pairs of $\langle$entity-name, entity-ID$\rangle$), they still struggle to obtain high-quality entity representations, which hinders their performance. Moreover, they follow the close world assumption, which assumes that the entity set is fixed and all entities appear during training. However, in practical scenarios, KBs evolve fast with novel entities added frequently \citep{dai2020inductively}. To represent novel entities, they need to retrain the model, which is extremely time-consuming.

% Despite their success, these approaches still leave much to be desired. Firstly, for existing entities, they are represented as indexes in the vocabulary, which lacks explicit modeling of entity semantic representations. Even with auxiliary pre-training and additional training data (pairs of $\langle$Entity-name, Entity-ID$\rangle$), they still struggle to obtain high-quality entity representations, which hinders their performance. Moreover, they follow the close world assumption, which assumes that the entity set is fixed and all entities appear during training. However, in practical scenarios, KBs evolve fast with novel entities added frequently \citep{dai2020inductively}. To represent novel entities, they need to retrain the model, which is extremely time-consuming.

% To address the above issues, we propose a novel framework, \textbf{B}i-\textbf{E}ncoder-\textbf{D}ecoder with \textbf{C}andidate \textbf{G}eneration. In contrast to the above methods that require labeled data which often suffers from sparsity, we take advantage of an entity encoder to represent entity by modeling their name and description which can be directly obtained in the KB. In this way, all entities are represented with shared parameters. Moreover, representations of newly added entities can be directly obtained without retraining, which is important in practice.

Inspired by bi-encoder models for entity retrival \citep{gillick2019learning, wu2020scalable}, we propose a novel framework, \textbf{B}i-\textbf{E}ncoder-\textbf{D}ecoder, to solve the above issues. Concretely, for existing entities, we take advantage of an entity encoder to obtain entity semantic representations by encoding entity names and descriptions. For novel entities, the entity representations can also be easily generated using the trained entity encoder without expensive retraining process. In addition, in order to improve the training and testing efficiency, an entity candidate set is generated for each sentence by comparing the surface-form similarity between entity name and the sentence.

% Additionally, we adopt BERT \citep{devlin2018bert} as an encoder to combine the power of CBED with the prior knowledge in pre-trained language models.

We evaluate our approach on the two benchmark datasets. Results from extensive experiments indicate that the proposed approach outperforms the state-of-the-art methods by a large margin and can handle novel entities well without retraining.

\section{Task Description}
Canonical relation extraction aims to extract relational triples from sentences whose elements (entity pairs and their relationship) are mapped to the knowledge base. This task can be defined as follows. Let $G = (E, R)$ be an existing KB, where $E$ and $R$ are the set of entities and relations in KB. Given a sentence $S$ as input, we aim to extract the triples $O = \{o_1, o_2, ..., o_j\}$ expressed by it, where $o_j = \langle h_j, r_j, t_j \rangle, h_j, t_j \in E$, and $r_j \in R$. Table  \ref{example} illustrates the input and target output of this task.

\section{Method}

In this section, we introduce our approach in details. We first perform candidate generation for the given sentence and then adopt a bi-encoder-decoder to translate sentences into triples.

\subsection{Candidate Generation}

The goal of this step is to provide a list of possible entities and filter irrelevant entities, given the sentence. To extract all spans which are possible entity mentions, we use Stanza \citep{qi2020stanza}, a open-source NLP tool to perform named entity recognition (NER) and part-of-speech (POS). After obtaining all possible entity mention spans, we use BM25, a variant of TF-IDF to measure similarity between mention spans and entity names. Top-K (64 in our experiments) candidate entities are kept by BM25 score for a sentence. On the dev set, the coverage of the top-64 candidates is 98.16\%.

% The goal of this step is to provide a list of possible entities with high recall because the coverage of the candidates determines the upper bound of the model. To extract all possible entity mention spans, we use Stanza \citep{qi2020stanza}, a open-source NLP tool to perform named entity recognition (NER) and part-of-speech (POS). After obtaining all possible entity mention spans, we use BM25, a variant of TF-IDF to measure similarity between mention string and entity names. Top-K (64 in our experiments) candidate entities are kept by BM25 score for a sentence.

% On the dev set, the coverage of top-64 candidates achieves 98.16\% when using NER and POS together. If we use NER and POS separately, the coverage are 94.86\% and 95.92\% respectively. This indicates that we benefit from an integrated approach in terms of coverage.

\subsection{Bi-Encoder-Decoder}

\begin{figure*}[t]
\centering
\includegraphics[width=0.8\textwidth]{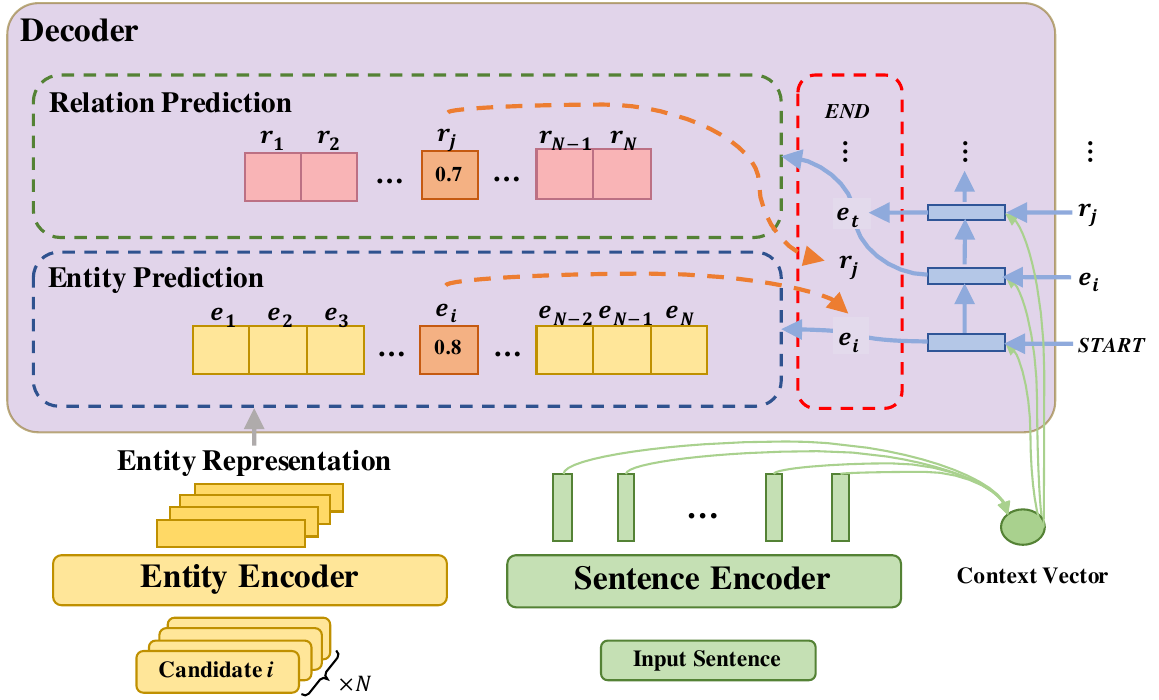} 
\caption{Bi-encoder-decoder model.}
\label{model}
\end{figure*}

Figure \ref{model} shows the overall framework of bi-encoder-decoder which consists of an entity encoder, a sentence encoder and a decoder.

\subsubsection{Entity Encoder}

The entity encoder is adopted to encode textual information of entities. Firstly, we obtain entity names and their descriptions which are off-the-self in the KB (Wikidata in our experiments). Then, the concatenation of entity name and description is fed to the entity encoder to generate the entity representation. Concretely, for each candidate entity, the input $\tau_{e}$ to our entity encoder is: (name [SEP] description), where name, description are words of entity name and description, and [SEP] is a special word to separate entity name and description. We employ GRU \citep{cho2014learning} as the encoder since it preforms similarly to LSTM \citep{hochreiter1997long}, but is computationally cheaper.

The encoding of entity information can be formulated as:
\begin{equation}
    \mathbf{H}_{e} = \textup{GRU}(\tau_{e}),
\end{equation}
\begin{equation}
    \mathbf{h}_{e} = \textup{MeanPooling}(\mathbf{H}_{e}),
\end{equation}
where $\mathbf{H}_{e}$ is the output of the hidden layer and the entity representation $\mathbf{h}_{e}$ is obtained by mean-pooling $\mathbf{H}_{e}$.

\subsubsection{Sentence Encoder}

We also adopt an encoder to model the representation of the input sentence. Similar with the entity encoder, a GRU network is employed to capture the contextual information of each word. Given a sentence $S = \langle w_1, w_2, ..., w_n \rangle$, for each word $w_i$, we use the hidden state of GRU as its representation.

\subsubsection{Decoder}

The decoder generates entity ID or relation ID at a time step and stops when a special token \textit{END} is generated. We also adopt GRU network as the decoder. At time step $t$, the decoder takes the source sentence encoding $\langle \mathbf{h}_1^s, \mathbf{h}_2^s, ..., \mathbf{h}_n^s \rangle$ and the previous target token embedding as input, and generates the hidden representation $\mathbf{h}_t^d$ for the current time step. Following previous work \citep{trisedya2019neural}, we also use a n-gram based attention mechanism to capture the multi-word entity mentions. We use $\mathbf{h}_t^d$ to make entity and relation prediction. The decoder switches to entity prediction mode and relation prediction mode in different time steps.

In the entity prediction mode, we assign each candidate entity a score by comparing its representation $\mathbf{h}_{e_i}$ with the current hidden representation $\mathbf{h}_t^d$. Specifically, we use dot-product to compute their semantic similarity and normalize the scores to obtain the probabilistic distribution:
\begin{equation}
    s_i = \mathbf{h}_t^d \cdot \mathbf{h}_{e_i},
\end{equation}
\begin{equation}
    % p(c_i | \mathbf{h}_t^D) = \frac{\textup{score}(\mathbf{h}_t^D, c_i)}{\sum_{i=1}^n{\textup{score}(\mathbf{h}_t^D, c_i)}},
    p(c_i | \mathbf{h}_t^d) = \frac{\exp(s_i)}{\sum_{j=1}^N \exp(s_j)},
\end{equation}
where $N$ is the number of candidate entities.

In the relation prediction mode, the current hidden representation $\mathbf{h}_t^d$ is projected to the predefined relation set $R$ using a linear layer.

\subsection{Training and Inference}

During training, ground-truth entities are added into the candidate entity set if they are missing. We train our model based on the cross-entropy loss which can be defined as follows: 
\begin{equation}
    \mathcal{L} = -\frac{1}{T} \sum_{t=1}^T \log {p(y_t|S)}.
\end{equation}
where $S$ is the input sentence and $y_t$ is the target token at time step $t$. During inference, generation is accomplished by searching over output sequences greedily while the training is teaching force.

\begin{table*}[t]
\centering

\resizebox{0.9\textwidth}{!}{
\begin{tabular}{l|c c c|c c c}   
\hline
\multirow{2}{*}{Models}&
\multicolumn{3}{c|}{WIKI}&
\multicolumn{3}{c}{GEO} \\
\multicolumn{1}{c|}{} & P & R & F1 & P & R & F1 \\

\hline
AIDA + CNN & 0.4035 & 0.3503 & 0.3750 & 0.3715 & 0.3165 & 0.3418 \\
NeuralEL + CNN & 0.3689 & 0.3521 & 0.3603 & 0.3781 & 0.3005 & 0.3349 \\
% \hline
% ClausIE + AIDA & 0.3617 & 0.4728 & 0.4099 & 0.3531 & 0.3951 & 0.3729 & - & - & -\\
% ClausIE + NeuralEL & 0.3445 & 0.3786 & 0.3607 & 0.3563 & 0.3791 & 0.3673 & - & - & -\\
%  \hline
AIDA + MinIE  & 0.3672 & 0.4856 & 0.4182 & 0.3574 & 0.3901 & 0.3730 \\
NeuralEL + MinIE & 0.3511 & 0.3967 & 0.3725 & 0.3644 & 0.3811 & 0.3726 \\
\hline
N-gram Attention & 0.7014 & 0.6432 & 0.6710 & 0.6029 & 0.6033 & 0.6031 \\
N-gram Attention (+pre-trained) & 0.7157 & 0.6634 & 0.6886 & 0.6581 & 0.6631 & 0.6606\\
N-gram Attention (+beam) & 0.7424 & \underline{0.6845} & 0.7123 & 0.6816 & \underline{0.6861} & 0.6838 \\
N-gram Attention (+triple classifier) & \textbf{0.8471} & 0.6762 & \underline{0.7521} & \textbf{0.7705} & 0.6771 & \underline{0.7208}\\

% \hline
% N-gram Attention$_{Bert}$ & \textbf{0.8899} & 0.7452 & 0.8112 & \textbf{0.9283} & \textbf{0.8940} & \textbf{0.9109} & 0.6508 & 0.5256 & 0.5816\\

\hline
Ours & \underline{0.8410} & \textbf{0.8264} & \textbf{0.8336} & \underline{0.7622} & \textbf{0.7660} & \textbf{0.7641}\\
% T-CRE$_{Bert}$ & 0.8762 & \textbf{0.8715} & \textbf{0.8739} & 0.8842 & 0.8860 & 0.8851 & \textbf{0.6708} & \textbf{0.6794} & \textbf{0.6752}\\

\hline
\end{tabular}
}

\caption{Main experimental results (\%). Numbers in bold mean the best results among all methods and the underlined ones mean the second best. For the compared methods, we use the reported results.}
\label{results}
\end{table*}

% \begin{table}[t]
% \centering
% \resizebox{.95\columnwidth}{!}{
% \begin{tabular}{l | r r r r}
% \hline
% Datasets & \#pairs & \#triples & \#entities & \#relations \\
% \hline
% Train & 224,881 & 225,729 & 248,244 & 157 \\

% Dev & 988 & 992 & 1,683 & 37 \\

% Test (WIKI) & 29,785 & 29,896 & 38,690 & 109 \\

% Test (GEO) & 1,000 & 1,000 & 124 & 11 \\
% \hline
% \end{tabular}
% }
% \caption{Statistics of the datasets.}
% \label{statistics}
% \end{table}

\section{Experiments}

% \begin{table*}[t]
% \centering

% \resizebox{\textwidth}{!}{
% \begin{tabular}{l|c c c|c c c c c c}   
% \hline
% \multirow{2}{*}{Models}&
% \multicolumn{3}{c|}{WIKI}&
% \multicolumn{6}{c}{GEO} \\
% \multicolumn{1}{c|}{} & P & R & F1 & P & R & F1 & Ign P & Ign R & Ign F1\\

% \hline
% N-gram Attention$_{Gru}$ (+triple classifier) & \textbf{0.8471} & 0.6762 & 0.7521 & 0.7705 & 0.6771 & 0.7208 & 0.2600$^*$ & 0.1666$^*$ & 0.2031$^*$\\
% \hline
% T-CRE$_{Gru}$ & 0.8435 & \textbf{0.8386} & \textbf{0.8410} & \textbf{0.8056} & \textbf{0.8120} & \textbf{0.8088} & \textbf{0.5000} & \textbf{0.5128} & \textbf{0.5223}\\

% \hline
% N-gram Attention$_{Bert}$ & 0.7932 & 0.7480 & 0.7700 & 0.8976 & 0.8940 & 0.8958 & 0.4286 & 0.4231 & 0.4258\\
% N-gram Attention$_{Bert}$ (+beam) & 0.8281 & 0.7798 & 0.8032 & 0.9055 & 0.9010 & 0.9033 & 0.5789 & 0.5641 & 0.5714\\
% N-gram Attention$_{Bert}$ (+triple classifier) & \textbf{0.8899} & 0.7452 & 0.8112 & \textbf{0.9283} & \textbf{0.8940} & \textbf{0.9109} & 0.6508 & 0.5256 & 0.5816\\
% \hline
% T-CRE$_{Bert}$ & 0.8762 & \textbf{0.8715} & \textbf{0.8739} & 0.8842 & 0.8860 & 0.8851 & \textbf{0.6708} & \textbf{0.6794} & \textbf{0.6752}\\

% \hline
% \end{tabular}
% }

% \caption{The effect of the encoder.}
% \label{encoder results}
% \end{table*}

\subsection{Datasets and Metrics}

We evaluate our model on two benchmark datasets \citep{trisedya2019neural}. They use distant supervision \citep{mintz2009distant} to align sentences in Wikipedia with relational triples in Wikidata. The collected dataset is splited into train set, dev set and test set. This test set is called as \textbf{WIKI}. For stress testing, they also collect another test set \textbf{GEO} from a travel website, which expresses in different style than Wikipedia. Following \citep{trisedya2019neural}, we use the metrics precision, recall and F1-score to measure the performance of different methods. 
% A predicted triple is deemed to be correct on the condition that the subject entity, object entity and their relation are all the same as those of the golden triple. 

% \subsection{Implementation Details}

% We adopt mini-batch mechanism to train our model with batch size 64; the learning rate is set to 2e-4 for both Gru encoder and decoder, and 2e-5 for Bert encoder. We use the same encoder for the input sentence and the entity information. We stop the training process when the performance on dev set does not gain any improvement in terms of F1-score. When using Gru encoder, we initialize word vectors with 512-dimension word embeddings by a uniform distribution $\mathcal{U}(-1,1)$. The dimension of Gru hidden state is set to 512. The pre-trained Bert encoder we used is [BERT-Base, Uncased], of which the dimension of hidden state is 768. The dimension of hidden state of Gru decoder is 512 and 768 for Gru encoder and Bert encoder respectively. All parameters are trainable during training and the optimizer we adopted is Adam \citep{kingma2014adam}.

\subsection{Compared Methods}

We compare our approach with the following methods: the combinations of two entity linking systems (\textbf{AIDA} \citep{hoffart2011robust} and \textbf{NeuralEL} \citep{kolitsas2018end}) and two relation extraction models (\textbf{CNN} \citep{lin2016neural} and \textbf{MiniE} \citep{gashteovski2017minie}); \textbf{N-gram Attention} \citep{trisedya2019neural} which is based on the encoder-decoder architecture and n-gram based attention mechanism; three enhanced variants of \textbf{N-gram Attention} using pre-trained embeddings, beam search and a triple classifier.

\subsection{Main Results}

The main experimental results are shown in Table \ref{results}. Compared to pipeline methods, \textbf{N-gram Attention} and its variants achieve promising results by avoiding cascading errors. The pre-trained embeddings provide a slight improvement and the beam search strategy and triple classifier significantly boost the performance of the model.
% Enhanced with the beam search strategy which re-rank the top-100 tail entities based on their string similarity with the sentence also boosts the performance of the model. Furthermore, a post-processing step is performed to filter invalid triples using the triple classifier, which obtain nearly 10\% improvement of precision.

It can be observed that our approach outperforms all the compared methods on recall and F1-score and achieves encouraging 8.15\% and 4.33\% improvements on F1-score over the previous state-of-the-art method \textbf{N-gram Attention (+triple classifier)} on \textbf{WIKI} and \textbf{GEO} respectively. Even without the triple classifier, our approach achieves competitive performance on precision. These observations demonstrate that our approach can effectively model the entities, which is important for CRE.

It is important to note that our model reduces the number of model parameters by a large margin and achieves significant inference speedup. Concretely, our model have about 39.3 million parameters, only one-fifth of the number of parameters in \textbf{N-gram Attention}. For inference, by filtering irrelevant entities in advance and performing a much simpler decoding strategy, the average inference time for each sentence can be reduced to about 30ms on a NVIDIA TESLA V100 GPU with nearly 5 times faster than the previous state-of-the-art.

\subsection{Results on novel Entities}

\begin{table}[t]
\centering
\resizebox{.95\columnwidth}{!}{
\begin{tabular}{p{3.2cm} | c c c}
\hline
Models & P & R & F1 \\
\hline

N-gram Attention & 0.6555 & 0.5855 & 0.6185 \\
N-gram Attention (+beam) & 0.7239 & \underline{0.6465} & 0.6831 \\
N-gram Attention (+triple classifier) & \underline{0.8117} & 0.6240 & \underline{0.7056} \\
\hline
Ours & \textbf{0.8401} & \textbf{0.8102} & \textbf{0.8249} \\
\hline

% N-gram Attention & 0.6785 & 0.6276 & 0.6521 \\
% N-gram Attention (+beam) & 0.7423 & 0.6873 & 0.7137 \\
% N-gram Attention (+triple classifier) & \textbf{0.8528} & 0.5985 & 0.7034 \\
% \hline
% BED & \underline{0.8482} & \textbf{0.8247} & \textbf{0.8363} \\
% \hline
\end{tabular}
}
\caption{Experiment results on \textbf{NEW}.}
\label{new results}
\end{table}

To investigate the models' ability to handle novel entites, we conduct the following experiment. To simulate the novel of new entities in the KB, we sample 1,000 entities from the whole entity set as novel entities which only appear in \textbf{WIKI}. The instances containing novel entities are selected from \textbf{WIKI} as a new test set \textbf{NEW}, which is used to evaluate the ability of handling novel entities.

% To investigate the models' ability to handle novel entites, we conduct the following experiments. To simulate the novel of new entities in the KB, we sample 1,000 entities which only appear in \textbf{WIKI} test set as newly added entities, denoted as $A$. The difference set of $E$ and $A$ can be denoted as $B$. The instances corresponding to new novel entities are filtered out from \textbf{WIKI} test set and denoted as \textbf{NEW} test set while the left instances are denoted as \textbf{OLD} test set.  

Assuming that we have trained models with existing entities, when new entities appear, we use these models to make predictions with or without change. For \textbf{N-gram Attention} and its variants, in order to get representations of novel entities, we expand the entity set, add pairs of $\langle$Entity-name, Entity-ID$\rangle$ into the train set and retrain these models. For our BED model, we directly use the previously trained model to make predictions on \textbf{NEW}, because the representations of novel entities can be easily generated using the off-the-shelf entity encoder.

% For \textbf{N-gram Attention} and its variants, we rebuild the entity vocabulary and retrain the models. N-gram Attention model and T-CRE model are trained using rebuilded target vocabulary and regenerated candidates, respectively. When new entities are novel, the N-gram Attention model should be rerun and then tested on \textbf{NEW} test set while our T-CRE model are directly tested without retraining.

% Next, we rebuild target entity vocabulary for N-gram Attention model and regenerate candidates for train, dev and \textbf{OLD} set. N-gram Attention model and T-CRE model are trained using rebuilded target vocabulary and regenerated candidates, respectively. When new entities are novel, the N-gram Attention model should be rerun and then tested on \textbf{NEW} test set while our T-CRE model are directly tested without retraining.

Table \ref{new results} reports the experimental results. We can observe that our model outperforms \textbf{N-gram Attention} and its variants on all three metrics. The results demonstrate that our proposed approach can handle novel entities well without retraining.

\section{Conclusion}

In this paper, we present BED, a novel framework for canonical relation extraction. To tackle the problems in existing methods based on the encoder-decoder architecture, we adopt an entity encoder to encode entity names and descriptions, which not only generates high-quality representations for existing entities, but also helps to model novel entities. Results on two benchmark datasets indicate that our approach performs better than the previous start-of-the-art methods and can handle novel entities well without retraining.

% In this paper, we present BED, a two-stage approach for canonicalized relation extraction. To tackle the problem of inefficient training over the whole entity set, we generate candidate entities for the input sentence, which provide more useful information to the model. To model the representation of candidates, we independently bootstrap entity representation from their name and description. Then, an encoder-decoder model with N-gram based attention mechanism is adopted to translate the input sentence into triples. In addition, we also integrate the powerful pre-trained language model.  futher improve T-CRE. Results on two benchmark datasets indicate that our approach achieves better performance than the previous start-of-the-art methods. Extensive experiment also demonstrates that our model is capable of handling new entities novel in the KB without retraining.

\bibliography{anthology,references}
\bibliographystyle{acl_natbib}

% \appendix

% \section{Example Appendix}
% \label{sec:appendix}

% This is an appendix.

\end{document}